\documentclass[10pt,twocolumn,letterpaper]{article}
\usepackage{preprint-layout}
\usepackage{multirow,array,placeins,float}
\newcommand{\tablecite}[1]{~\cite{#1}}
\definecolor{linkblue}{rgb}{0.21,0.49,0.74}
\usepackage[breaklinks,colorlinks,allcolors=linkblue]{hyperref}
\pdftrailerid{}
\hypersetup{pdftitle={Let Each Modality Play to Its Strengths in RGB–Event Moving Object Detection},pdfauthor={Ziheng Wang, Chaolang Li, Yutong Yang, Xiaohan Xu, Chongxiang Yang, Yue Chen, Hengxuan Zhong, Zhen Liang, Pengwen Dai},pdfcreator={},pdfproducer={}}
\title{Let Each Modality Play to Its Strengths\\in RGB--Event Moving Object Detection}
\author{%
  Ziheng Wang$^{1}$ \quad Chaolang Li$^{1}$ \quad Yutong Yang$^{1}$ \quad Xiaohan Xu$^{1}$ \quad Chongxiang Yang$^{1}$\\[2pt]
  Yue Chen$^{1}$ \quad Hengxuan Zhong$^{1}$ \quad Zhen Liang$^{2}$ \quad Pengwen Dai$^{1,*}$\\[5pt]
  {\small $^{1}$Sun Yat-sen University \quad $^{2}$Shenzhen University}\\[3pt]
  {\small $^{*}$Corresponding author: \href{mailto:daipw@mail.sysu.edu.cn}{\texttt{daipw@mail.sysu.edu.cn}}}%
}

\begin{document}
\maketitle
\begin{abstract}
State-of-the-art RGB--Event detectors improve the detection of fast-moving objects by combining complementary RGB and Event features, yet they typically fuse the two modalities into a unified representation for both localization and classification. We propose an \textbf{A}symmetric \textbf{E}vent--\textbf{R}GB \textbf{O}bject \textbf{Det}ection Transformer (\textbf{AERODet}), which allows localization and classification to adopt different fusion strategies. For localization, we design Supervised Unimodal Response Exploitation (SURE), which independently supervises each modality branch and exploits their responses to calibrate multimodal fusion weights. For classification, we introduce Disentangled Semantic and Objectness Calibration (DSOC), which decouples classification from localization, injects joint RGB--Event semantic features for category prediction, and further exploits the Event RoI responses of SURE to calibrate objectness. The combination of these two components enables each modality to fully play to its strengths. Experiments on FRED and NeRDD demonstrate that AERODet achieves state-of-the-art performance, surpassing the strongest RGB--Event baselines on FRED by 7.7 AP50 points and 4.6 mAP points.
\end{abstract}
\section{Introduction}
\label{sec:intro}

Moving object detection is an important computer vision task for airspace monitoring, security, and robotic perception, yet it remains challenging due to long-range observation, small target size, and rapid motion~\cite{xu2022rfla,magrini2025droneSurvey}. Conventional frame-based RGB sensors struggle under rapid motion and are highly sensitive to lighting conditions, as their limited temporal resolution causes motion blur and their low dynamic range leads to overexposure or underexposure, severely degrading target visibility~\cite{magrini2025droneSurvey,magrini2025fred}. Event cameras alleviate these limitations by asynchronously recording per-pixel brightness changes with high temporal resolution and high dynamic range, thereby preserving motion-sensitive cues for fast-moving targets~\cite{perot2020red,gallego2022event}. However, Event representations lack fine-grained appearance cues such as color and texture, which are important for classification. RGB--Event detection can therefore combine the strengths of Event data with the rich appearance of RGB images, offering a promising solution for robust moving object detection under challenging conditions~\cite{gehrig2024dagr,magrini2024nerdd,magrini2025fred}.

\begin{figure}[t]
\centering
\includegraphics[width=\columnwidth]{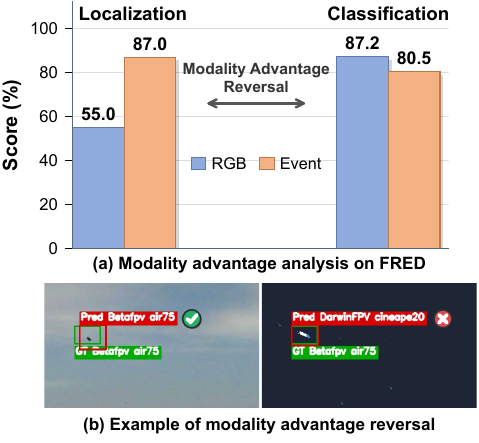}
\caption{Influence of the two modalities on localization and classification. RGB yields a looser box with the correct category, whereas Event yields a tighter box with an incorrect category.}
\label{fig:teaser}
\end{figure}

Despite this complementarity, existing RGB--Event detectors mainly focus on task-symmetric fusion architectures~\cite{tomy2022fpnfusion,zhou2023renet,cao2024eolo,cao2024cafr,lu2025flexevent}. For instance, SODFormer~\cite{li2023sodformer} models temporal dependencies and cross-modal interaction with a streaming Transformer, while SPFD~\cite{wang2026spfd} separates shared and modality-specific features to preserve complementary information. However, these methods ultimately rely on the same fused features for both localization and classification, without considering the distinct strengths of RGB and Event across different tasks. Moreover, when both modalities are trained only through the fused prediction, the dominant modality may suppress the learning of the other, leaving both modalities insufficiently optimized~\cite{wang2020gradientblending,peng2022ogm,wu2022greedy,zhao2025m2dlif}. These limitations motivate us to re-examine the rationality of such task-symmetric architectures.

To investigate this issue, we conduct a modality-specific exploration using independently trained RGB and Event detectors on FRED~\cite{magrini2025fred}. As shown in Figure~\ref{fig:teaser}, Event provides much stronger localization, whereas RGB achieves higher classification accuracy. These results indicate that Event is more effective for localizing fast-moving objects, while RGB provides stronger category evidence. We also find that this modality advantage gap is much larger than that in RGB--Infrared detection, as shown in Table~\ref{tab:task-analysis}. Therefore, the task-symmetric fusion widely adopted in RGB--Infrared detection~\cite{fang2021cft,shen2024icafusion,yuan2024c2former,guo2024damsdet} may not fully exploit the strengths of RGB--Event data, which calls for a task-asymmetric design that allows localization and classification to adopt different fusion strategies.

Based on this observation, we propose an end-to-end \textbf{A}symmetric \textbf{E}vent--\textbf{R}GB \textbf{O}bject \textbf{Det}ection Transformer (\textbf{AERODet}) to tailor fusion strategies to both subtasks. Specifically, we propose Supervised Unimodal Response Exploitation (SURE) for localization, which uses a supervised detection head independent of the main detection path to sufficiently optimize each modality branch. It further estimates modality reliability from the head responses and calibrates the decoder attention weights. However, since the Event modality dominates localization for moving objects, the decoder attention weights may be biased toward Event, causing the shared queries to lose valuable RGB category cues. For classification, we therefore introduce Disentangled Semantic and Objectness Calibration (DSOC) to decouple each final query into localization and objectness queries. From the boxes of the localization query, DSOC injects joint RGB--Event semantic features into the objectness query for category prediction, while calibrating its objectness score with the Event RoI response of SURE. Together, SURE and DSOC allow both subtasks to fully exploit the complementary strengths of each modality.

Our contributions are summarized as follows:
\begin{itemize}
\item We rethink task-symmetric architectures in RGB--Event moving object detection, revealing a task-dependent modality advantage reversal and exploring an asymmetric paradigm that lets each modality play to its strengths.

\item We propose SURE for localization and DSOC for classification, calibrating attention weights via supervised unimodal responses and injecting joint RGB--Event semantic features with Event-guided objectness calibration.

\item Extensive experiments on FRED and NeRDD show that AERODet achieves state-of-the-art performance across benchmarks, data splits, and evaluation settings, validating the effectiveness of our proposed paradigm.
\end{itemize}
\section{Related Work}
\label{sec:related}

\subsection{Object Detection}

Object detection jointly addresses localization and classification. Conventional detectors either classify and refine region proposals~\cite{girshick2014rcnn,girshick2015fastrcnn,ren2015fasterrcnn,lin2017fpn} or directly predict boxes and categories in one pass~\cite{redmon2016yolo,tian2019fcos,zhou2019objects,khanam2024yolov11,tian2025yolov12}, while DETR-based detectors formulate detection as set prediction with object queries~\cite{carion2020detr,zhu2021deformabledetr,meng2021conditionaldetr,liu2022dabdetr,li2022dndetr,zhang2023dino,zhang2023ddq,zong2023codetr,zhao2024rtdetr}. Among them, RT-DETR~\cite{zhao2024rtdetr} further enables real-time end-to-end detection with an efficient hybrid encoder. Since the two tasks rely on different evidence, previous studies separate their prediction heads, disentangle their spatial features, or align their predictions~\cite{wu2020doublehead,song2020tsd,li2020gfl,feng2021tood}. Decoupled DETR~\cite{zhang2023decoupleddetr} observes that classification favors salient regions while localization benefits from boundaries, and introduces task-aware queries and separate cross-attention. However, these methods are built on a single image modality. In contrast, AERODet decouples how the two tasks exploit RGB and Event evidence.

Event-based object detectors mainly differ in how they represent and process event streams. Some methods preserve the sparse spatiotemporal structure of events~\cite{schaefer2022aegnn,li2022astmnet,sadoun2026sparsevoxel}, whereas others aggregate them into grid-like tensors for CNN, Transformer, or state-space detectors~\cite{perot2020red,gehrig2023rvt,peng2024sast,zubic2024ssm,yang2025smamba}. With the high temporal resolution and dynamic range of Event data, these approaches can effectively detect fast-moving objects~\cite{magrini2025droneSurvey}. However, Event representations provide limited appearance cues, such as color and texture, and may become sparse under weak relative motion. These limitations motivate combining Event motion evidence with rich RGB appearance for moving object detection.

\begin{figure*}[t]
\centering
\includegraphics[width=\textwidth]{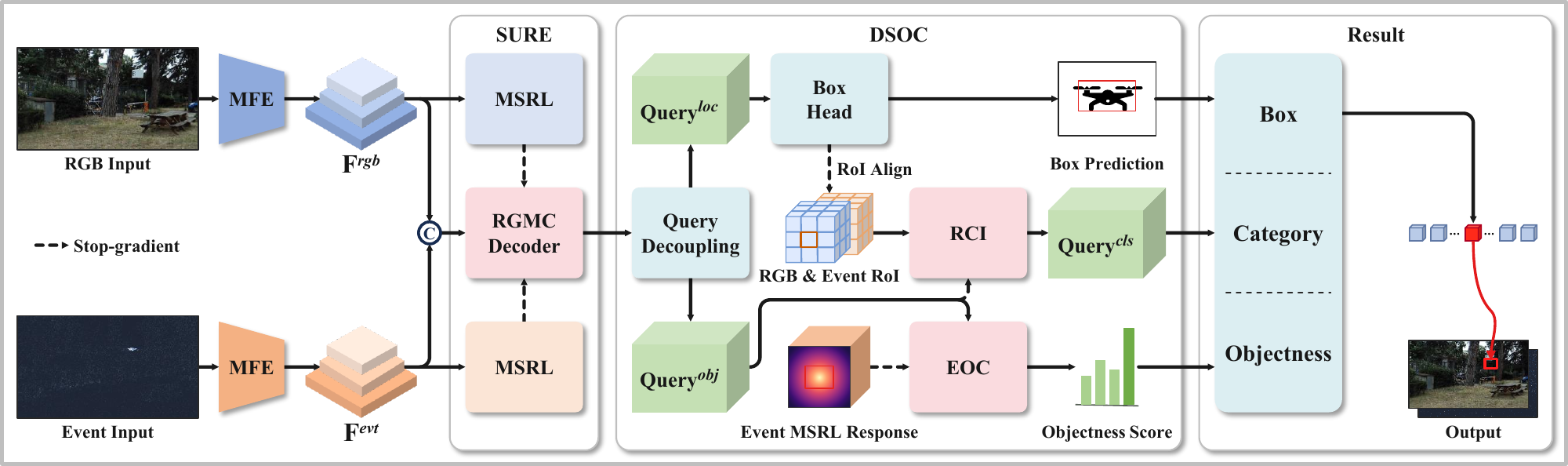}
\caption{Overview of AERODet. In Supervised Unimodal Response Exploitation (SURE), Modality-Specific Response Learning (MSRL) supervises each modality branch, and Response-Guided Modality Calibration (RGMC) calibrates decoder attention weights. In Disentangled Semantic and Objectness Calibration (DSOC), Regional Category Injection (RCI) injects joint RoI features into the objectness query, and Event-guided Objectness Calibration (EOC) calibrates objectness with the Event RoI response. AERODet is trained end-to-end.}
\label{fig:framework}
\end{figure*}
\subsection{Multimodal Learning and Fusion}

Multimodal learning often suffers from modality imbalance. When all modalities are optimized toward a joint objective, they tend to overfit and generalize at different rates~\cite{wang2020gradientblending,wu2022greedy}, and the dominant modality can suppress the optimization of the others~\cite{peng2022ogm,fan2023pmr}. Accordingly, previous methods reweight the modality-specific losses~\cite{wang2020gradientblending}, modulate the gradient of each modality~\cite{peng2022ogm}, guide the under-optimized modality with prototypes~\cite{fan2023pmr}, integrate conflicting gradients in a Pareto manner~\cite{wei2024mmpareto}, or weight modalities by data quality~\cite{zhang2023qmf}. ARL~\cite{wei2025arl} takes a different view, arguing that equal dependence on each modality is not necessarily optimal, and adjusts the modality contributions by their prediction variance. However, these methods mainly target multimodal classification with a single objective, whereas the localization and classification subtasks of detection may favor different modality evidence.

Multimodal object detectors combine RGB images with complementary modalities to improve robustness under challenging conditions~\cite{zhang2019arcnn,zhou2020mbnet,sun2022uacmdet}. In RGB--Infrared detection, CFT~\cite{fang2021cft} models cross-modal interaction with Transformers, ICAFusion~\cite{shen2024icafusion} and C$^2$Former~\cite{yuan2024c2former} refine complementary features via cross-attention, DAMSDet~\cite{guo2024damsdet} selects competitive queries from both modalities, DyFCLT~\cite{li2026dyfclt} performs frequency-decoupled cross-modal learning, and M$^2$D-LIF~\cite{zhao2025m2dlif} distills unimodal features from pre-trained teachers to alleviate insufficient mono-modality learning. Despite their effectiveness in low light, frame-based infrared sensors still suffer from motion blur on fast-moving targets. To overcome this, recent studies turn to high-temporal-resolution Event sensors~\cite{tomy2022fpnfusion,gehrig2024dagr,cao2024cafr}, developing RGB--Event detectors such as RENet~\cite{zhou2023renet}, EOLO~\cite{cao2024eolo}, and FlexEvent~\cite{lu2025flexevent} to fuse both modalities under rapid motion and varying illumination. However, these methods still share the same modality weights for localization and classification. In contrast, AERODet reuses the responses of independently supervised branches and allows both tasks to combine the two modalities with separate weights and features, thereby promoting balanced modality optimization while enabling each modality to play to its strengths.
\section{Method}
\label{sec:method}

\subsection{Overall Framework}
As illustrated in Figure~\ref{fig:framework}, AERODet consists of Multi-scale Feature Extraction (MFE), Supervised Unimodal Response Exploitation (SURE), and Disentangled Semantic and Objectness Calibration (DSOC). Given RGB and Event images $I^{rgb}$ and $I^{evt}$, MFE first extracts three-scale features $\{\mathbf F_l^{rgb}\}_{l=1}^{3}$ and $\{\mathbf F_l^{evt}\}_{l=1}^{3}$ using two modality-specific backbone and encoder branches following RT-DETR~\cite{zhao2024rtdetr}. The tokens with the highest confidence scores are selected from the six feature maps to initialize the object queries and reference boxes following DAMSDet~\cite{guo2024damsdet}. Then, SURE attaches a separately supervised Modality-Specific Response Learning (MSRL) head to each branch to sufficiently optimize both modality branches. It further uses the scale-wise reliabilities estimated from the detached objectness responses to modulate the RGB and Event attention weights in the decoder. Next, DSOC decouples the final query into localization and objectness queries, injects joint RGB--Event RoI features into the latter for category prediction, and calibrates the class-agnostic objectness logit with the Event RoI response. Finally, the detection scores are the products of the category and calibrated objectness scores.

\subsection{SURE}
\label{sec:sure}
\begin{figure}[t]
\centering
\includegraphics[width=\columnwidth]{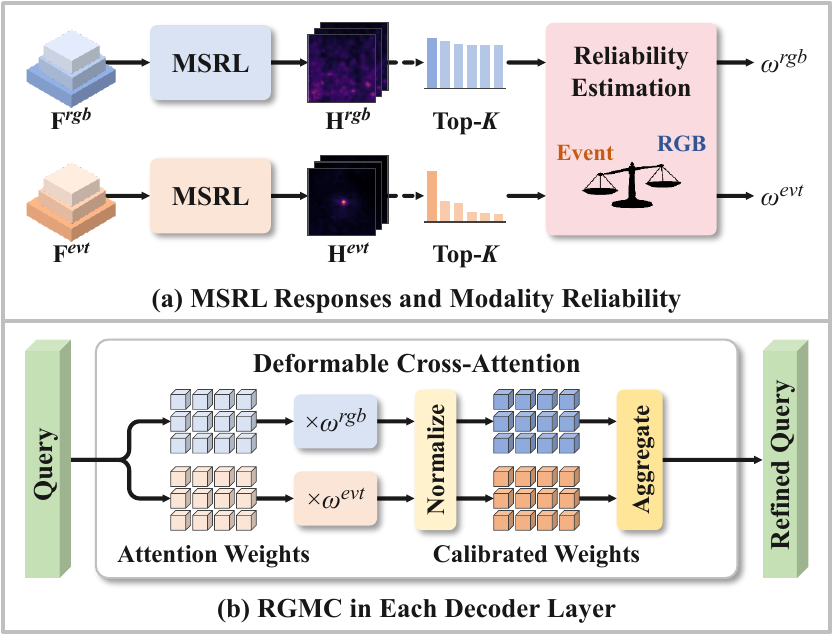}
\caption{Illustration of SURE. (a) The MSRL head of each modality produces objectness responses, from which the modality weights are estimated. (b) In each decoder layer, RGMC rescales the RGB and Event attention weights before joint normalization. Dashed arrows denote stop-gradient.}
\label{fig:sure}
\end{figure}
As illustrated in Figure~\ref{fig:sure}, SURE first learns modality-specific responses under separate supervision to alleviate the insufficient optimization of both modalities, and then exploits them to calibrate the RGB and Event attention weights at each decoder layer for localization.

\noindent\textbf{Modality-Specific Response Learning (MSRL).}
For each modality $m\in\{rgb,evt\}$, MSRL attaches a separate head that is shared across its three-scale encoder features. Given the feature map $\mathbf F$ of modality $m$ at scale $l$, the head generates the category heatmap $\hat{\mathbf Y}=\sigma(\Psi_{hm}(\mathbf F))\in\mathbb R^{H\times W\times C}$, the offset map $\hat{\mathbf O}=\Psi_{o}(\mathbf F)\in\mathbb R^{H\times W\times 2}$, and the size map $\hat{\mathbf S}=\Psi_{s}(\mathbf F)\in\mathbb R^{H\times W\times 4}$, where $\Psi_{hm}$, $\Psi_{o}$, and $\Psi_{s}$ are convolutional layers as in~\cite{zhou2019objects}, $C$ is the number of categories, and $\sigma(\cdot)$ is the sigmoid function. We regard object center localization as pixel-wise logistic regression with focal loss~\cite{lin2017focal,zhou2019objects}, which is defined as
\begin{equation}
\mathcal L_{hm}=-\frac{1}{N}\sum_{i}\!\left\{
\begin{array}{@{}l@{\,}l@{}}
(1-\hat Y_i)^{\gamma_1}\log\hat Y_i & Y_i{=}1,\\
(1-Y_i)^{\gamma_2}\hat Y_i^{\gamma_1}\log(1-\hat Y_i) & \text{o.w.,}
\end{array}\right.
\label{eq:focal}
\end{equation}
where $N$ is the number of objects, $i$ indexes the heatmap elements, $\mathbf Y$ is the ground-truth heatmap generated following~\cite{zhou2019objects}, and the penalty hyper-parameters $\gamma_1$ and $\gamma_2$ are set to 2 and 4. The offset and size regression only considers object centers, whose objectives are formulated as
\begin{equation}
\begin{aligned}
\mathcal L_{o}&=\frac{1}{2N}\sum_{k=1}^{N}\big\|\hat{\mathbf O}_{k}-\mathbf O_{k}\big\|_1,\\
\mathcal L_{s}&=\frac{1}{4N}\sum_{k=1}^{N}\big\|\hat{\mathbf S}_{k}-\mathbf S_{k}\big\|_1,
\end{aligned}
\end{equation}
where $\hat{\mathbf O}_k$ and $\hat{\mathbf S}_k$ are the predicted offset and size at the $k$-th object center with ground truths $\mathbf O_k$ and $\mathbf S_k$. The MSRL loss of modality $m$ sums the three losses over all scales,
\begin{equation}
\mathcal L_{\mathrm{msrl}}^m=\sum_{l=1}^{3}\big(\mathcal L_{hm}+\mathcal L_{o}+\lambda_{s}\mathcal L_{s}\big),
\label{eq:aux-loss}
\end{equation}
where the losses inside the sum are computed on the $l$-th feature map of modality $m$, and $\lambda_s$ is set to 0.1.

Since the head of modality $m$ only receives the features of this modality, $\mathcal L_{\mathrm{msrl}}^m$ directly optimizes its encoder and backbone, regardless of its contribution to the fused prediction. Thus, both modalities receive the same form of task supervision to promote sufficient optimization of each branch. Moreover, we take the channel-wise maximum of $\hat{\mathbf Y}$ as the objectness response $\mathbf H_l^m$, which is detached and reused by the following modules.

\noindent\textbf{Response-Guided Modality Calibration (RGMC).}
Given the sparse nature of Event data and the small number of foreground targets, reliable evidence is expected to produce a few dominant objectness responses. Moreover, inspired by ARL's use of predictive uncertainty to characterize modality reliability~\cite{wei2025arl}, MSRL detection responses can also serve as a proxy for predictive certainty. RGMC therefore retains the \(K\) largest values in each objectness response map $\mathbf H_l^m$, where $K=\max(1,\lfloor\eta HW\rfloor)$, $H\times W$ is the spatial size of $\mathbf H_l^m$, and $\eta$ is the retention ratio. Let \(t_j\) denote the \(j\)-th largest retained value. We quantify the concentration of the retained responses using peak dominance \(D\) and normalized entropy \(U\), which serve as indicators of predictive certainty and are defined as
\begin{equation}
D=\frac{t_1}{\bar t},\quad
U=-\frac{1}{\log K}\sum_{j=1}^{K}p_j\log p_j,
\end{equation}
where $\bar t=\sum_{j=1}^{K}t_j/K$ is the mean retained response and $p_j=t_j/(K\bar t)$ is the normalized response. A large $D$ indicates that the strongest response clearly dominates the others, while a large $U$ indicates evenly distributed and thus ambiguous responses. We then compute the relative reliability of modality $m$ at scale $l$ from $D_l^m$ and $U_l^m$ as
\begin{equation}
\alpha_l^m=\frac{D_l^m/U_l^m}{\sum_{m'}D_l^{m'}/U_l^{m'}}.
\label{eq:reliability}
\end{equation}
In each decoder layer, deformable cross-attention~\cite{zhu2021deformabledetr} aggregates the sampled RGB and Event features with weights normalized jointly over modalities, scales, and sampling points. RGMC multiplicatively modulates the weight $A_l^m$ of each sampling point from modality $m$ at scale $l$ as
\begin{equation}
\widetilde A_l^m=\frac{\omega_l^m A_l^m}{\sum_{m',l'}\omega_{l'}^{m'}A_{l'}^{m'}},\quad
\omega_l^m=\Big(\frac{\alpha_l^m}{\alpha_l^{\bar m}}\Big)^{\frac12},
\label{eq:bias}
\end{equation}
where $\bar m$ denotes the other modality and the sum runs over all sampling points of the same query and attention head. Thus, the Event-to-RGB weight ratio at scale $l$ is multiplied by $\alpha_l^{evt}/\alpha_l^{rgb}$, which shifts the aggregation toward the more reliable modality, while the sampling offsets and the relative weights within each modality--scale level remain unchanged, as detailed in the supplementary material.

\subsection{DSOC}
\label{sec:dsoc}
Although SURE calibrates the multimodal evidence for localization, the final query is still shared by box regression and category recognition, which favor different evidence. DSOC therefore decouples the final query to inject joint RGB--Event RoI features for semantic complementarity and calibrate objectness with the Event RoI response.

For a query of the final decoder layer, we denote its representation and reference box~\cite{liu2022dabdetr} by $\mathbf q$ and $\bar{\mathbf b}$. For each task $\tau\in\{loc,obj\}$, a residual adapter produces the task-specific query $\mathbf q^{\tau}$, which is formulated as
\begin{equation}
\mathbf q^{\tau}=\operatorname{LN}(\mathbf q+\operatorname{Adapter}_{\tau}(\mathbf q)),
\end{equation}
where $\operatorname{LN}(\cdot)$ denotes layer normalization. The localization head predicts a box residual from $\mathbf q^{loc}$ to refine $\bar{\mathbf b}$ in inverse-sigmoid space~\cite{zhu2021deformabledetr}, and the objectness head predicts an objectness logit from $\mathbf q^{obj}$, which are formulated as
\begin{equation}
\begin{aligned}
\hat{\mathbf b}&=\sigma\!\left(\sigma^{-1}(\bar{\mathbf b})+\operatorname{BBoxHead}(\mathbf q^{loc})\right),\\
\hat o&=\operatorname{ObjHead}(\mathbf q^{obj}),
\end{aligned}
\end{equation}
where $\sigma^{-1}(\cdot)$ denotes the inverse-sigmoid function.

\noindent\textbf{Regional Category Injection (RCI).}
Localization and classification may require different modality weights. However, the decoder weights are calibrated by SURE from class-agnostic objectness responses and shared by both subtasks, which may not suit category recognition. Therefore, once the final box $\hat{\mathbf b}$ is obtained, RCI revisits the regional features of both modalities to complement the category cues of the query. Specifically, RCI extracts the $3\times3$ RoIAlign feature~\cite{he2017maskrcnn} $\mathbf R^m$ from the first-scale feature map $\mathbf F_1^m$ of each modality $m\in\{rgb,evt\}$ and average-pools it into a regional token $\mathbf t^m$, which is formulated as
\begin{equation}
\begin{aligned}
\mathbf R^m&=\operatorname{RoIAlign}_{3\times3}(\operatorname{sg}(\mathbf F_1^m),\operatorname{sg}(\hat{\mathbf b})),\\
\mathbf t^m&=\operatorname{AvgPool}(\mathbf R^m),\\
\mathbf t^{joint}&=\operatorname{Proj}_{joint}(\operatorname{Concat}(\mathbf t^{rgb},\mathbf t^{evt})).
\end{aligned}
\label{eq:rci}
\end{equation}
Here, $\operatorname{sg}(\cdot)$ denotes stop-gradient, and $\operatorname{Proj}_{joint}$ is a linear layer that maps the concatenated tokens into the joint category token $\mathbf t^{joint}$. Thus, $\operatorname{Proj}_{joint}$ learns the modality weights for categorization, separately from the localization-oriented weights in the decoder. This token is then injected into the detached objectness query $\mathbf q^{obj}$ as a residual to obtain the classification query $\mathbf q^{cls}$,
\begin{equation}
\begin{aligned}
\mathbf q^{cls}&=\operatorname{LN}(\operatorname{sg}(\mathbf q^{obj})+\beta\mathbf t^{joint}),\\
\hat{\mathbf s}^{cls}&=\sigma(\operatorname{ClsHead}(\mathbf q^{cls})),
\end{aligned}
\end{equation}
where $\beta$ controls the contribution of the joint category token, and $\operatorname{ClsHead}$ denotes the classification head that predicts the scores $\hat{\mathbf s}^{cls}$ of the $C$ categories.

\noindent\textbf{Event-guided Objectness Calibration (EOC).}
Since Event cameras only respond to brightness changes, moving objects trigger events while the static background is largely suppressed. Thus, the Event objectness response learned by MSRL serves as a natural foreground indicator. EOC exploits this property by calibrating the objectness logit $\hat o_j$ of each query into $\widetilde o_j$ with its Event RoI response as
\begin{equation}
\begin{aligned}
r_j&=\max\operatorname{RoIAlign}_{3\times3}(\mathbf H_1^{evt},\hat{\mathbf b}_j),\\
\widetilde o_j&=\hat o_j+\rho\log\big(r_j/\bar r\big),
\end{aligned}
\label{eq:eoc}
\end{equation}
where $j$ indexes the decoder queries, $\mathbf H_1^{evt}$ is the Event response on the highest-resolution map $\mathbf F_1^{evt}$, $r_j$ is the peak Event response within the $j$-th predicted box $\hat{\mathbf b}_j$, $\bar r$ is the mean of $r_j$ over all queries, and $\rho$ controls the calibration strength. Thus, boxes covering stronger Event responses than the average receive higher objectness. Finally, the detection score $\hat s_{j,c}^{det}$ of the $j$-th query for category $c$ combines the calibrated objectness and the category score as
\begin{equation}
\hat s_{j,c}^{det}=\sigma(\widetilde o_j)\,\hat s_{j,c}^{cls}.
\end{equation}

\subsection{Training Objective}
AERODet is trained end-to-end with the total loss
\begin{equation}
\mathcal L=\mathcal L_{\mathrm{det}}
+\lambda_{\mathrm{msrl}}\big(\mathcal L_{\mathrm{msrl}}^{rgb}+\mathcal L_{\mathrm{msrl}}^{evt}\big)
+\lambda_{\mathrm{rci}}\mathcal L_{\mathrm{rci}},
\end{equation}
where the balance factors $\lambda_{\mathrm{msrl}}$ and $\lambda_{\mathrm{rci}}$ are set to 1.0 and 0.25. The main detection loss $\mathcal L_{\mathrm{det}}$ follows DINO~\cite{zhang2023dino} and supervises the class-agnostic objectness and box predictions, and $\mathcal L_{\mathrm{rci}}$ applies the Varifocal loss~\cite{zhang2021varifocal} to the category predictions of RCI with the matching assignments of the main detection branch. Owing to the stop-gradient operations, $\mathcal L_{\mathrm{det}}$ trains the decoder and both feature branches through the fused pathway, $\mathcal L_{\mathrm{msrl}}^{m}$ additionally supervises the branch of modality $m$, and $\mathcal L_{\mathrm{rci}}$ only updates RCI.
\section{Experiments}
\label{sec:experiments}
\subsection{Datasets and Implementation Details}
\noindent\textbf{Datasets.}
FRED~\cite{magrini2025fred} is a high-resolution RGB--Event drone detection benchmark with more than 7 hours of synchronized $1280\times720$ recordings of five drone models under diverse weather and illumination. We follow its canonical split with balanced recording conditions and its challenging split with modality-specific shifts caused by illumination, weather, and insects. NeRDD~\cite{magrini2024nerdd} contains more than 3.5 hours of synchronized $1280\times720$ RGB--Event recordings across diverse scenes and backgrounds. All compared methods use its video-level split, with 80\% of the videos for training and the rest for testing. For the modality advantage analysis, we also use the RGB--Infrared (IR) datasets M3FD~\cite{liu2022tar}, FLIR~\cite{zhang2020flir}, and VEDAI~\cite{razakarivony2016vedai} with six, three, and eleven categories, respectively.

\noindent\textbf{Evaluation protocols.}
Following the standard COCO protocol~\cite{lin2014coco}, we adopt AP50 and mAP for evaluation. The official FRED protocol treats all drones as one category, and NeRDD only provides single-category annotations. Since FRED also provides the labels of the five drone models, we further use them on the canonical split to study the complementarity of RGB and Event in category recognition. For the modality advantage analysis in Section~\ref{sec:task-analysis}, localization is measured by class-agnostic AP50 and mAP, and classification by Accuracy and Macro-F1~\cite{sokolova2009performance}.

\noindent\textbf{Implementation details.}
We employ COCO-pretrained ResNet-50 backbones~\cite{he2016resnet,lin2014coco} for the RGB and Event branches in MFE. In SURE, the response retention ratio $\eta$ is set to $0.01$. In DSOC, we set the joint token weight $\beta$ and Event calibration strength $\rho$ to $0.1$ and $0.5$, respectively. AERODet remains stable within the tested ranges of $\eta$, $\beta$, and $\rho$, as shown in the supplementary material. We train AERODet using AdamW~\cite{loshchilov2019adamw} with a learning rate of $1\times10^{-4}$, a batch size of 4, and standard data augmentations. For fair comparisons, we set the input size to $640\times640$ for testing. Other settings follow RT-DETR~\cite{zhao2024rtdetr} and are listed in the supplementary material.

\subsection{Task-Dependent Modality Strengths}
\label{sec:task-analysis}
\begin{table}[t]
\centering\small
\setlength{\tabcolsep}{2pt}
\caption{Task-dependent modality analysis.}
\label{tab:task-analysis}
\begin{tabular*}{\columnwidth}{@{\extracolsep{\fill}}lllrrrr}
\toprule
 & & & \multicolumn{2}{c}{Localization} & \multicolumn{2}{c}{Classification} \\
\cmidrule(lr){4-5}\cmidrule(lr){6-7}
Dataset & Probe & Modality & AP50 & mAP & Accuracy & Macro-F1 \\
\midrule
\multirow{4}{*}{FRED} & \multirow{2}{*}{MSRL} & RGB & 46.0 & 14.4 & \textbf{88.4} & \textbf{87.8} \\
 & & Event & \textbf{85.5} & \textbf{43.3} & 84.3 & 82.2 \\
\cmidrule(lr){2-7}
  & \multirow{2}{*}{RT-DETR} & RGB & 55.0 & 20.8 & \textbf{87.2} & \textbf{86.4} \\
 & & Event & \textbf{87.0} & \textbf{48.6} & 80.5 & 76.2 \\
\midrule
\multirow{2}{*}{M3FD} & \multirow{2}{*}{RT-DETR} & RGB & 88.1 & \textbf{58.2} & 94.5 & 85.1 \\
 & & IR & \textbf{88.3} & 56.8 & \textbf{96.8} & \textbf{89.2} \\
\midrule
\multirow{2}{*}{FLIR} & \multirow{2}{*}{RT-DETR} & RGB & 80.1 & 44.2 & 97.8 & 96.0 \\
 & & IR & \textbf{88.5} & \textbf{54.3} & \textbf{98.7} & \textbf{96.5} \\
\midrule
\multirow{2}{*}{VEDAI} & \multirow{2}{*}{RT-DETR} & RGB & \textbf{92.5} & 56.1 & \textbf{74.6} & \textbf{62.3} \\
 & & IR & 92.4 & \textbf{58.0} & 72.7 & 60.3 \\
\bottomrule
\end{tabular*}
\end{table}
\begin{table}[t]
\centering\small
\setlength{\tabcolsep}{2pt}
\caption{Cross-task recombination on FRED five-class. RT-DETR values are diagnostic AP on GT-matched queries.}
\label{tab:fred-recombination}
\begin{tabular*}{\columnwidth}{@{\extracolsep{\fill}}lllrr}
\toprule
Probe & Localization & Category & AP50 & mAP \\
\midrule
\multirow{4}{*}{MSRL} & RGB & RGB & 40.3 & 12.3 \\
 & RGB & Event & 41.8 & 12.9 \\
 & Event & Event & 66.4 & 31.8 \\
 & Event & RGB & \textbf{71.9} & \textbf{34.3} \\
\midrule
\multirow{4}{*}{RT-DETR} & RGB & RGB & 52.4 & 19.3 \\
 & RGB & Event & 45.1 & 17.4 \\
 & Event & Event & 63.4 & 38.0 \\
 & Event & RGB & \textbf{75.6} & \textbf{44.1} \\
\bottomrule
\end{tabular*}
\end{table}
To avoid presupposing the roles of the two modalities, we probe them with the MSRL heads of a task-symmetric baseline trained without RGMC and DSOC, and with five-class RGB-only and Event-only RT-DETRs~\cite{zhao2024rtdetr}. Classification is evaluated within the same ground-truth (GT) boxes for the MSRL heads and on GT-matched queries for the RT-DETRs, as detailed in the supplementary material.

\noindent\textbf{Modality preference on FRED.}
As shown in Table~\ref{tab:task-analysis}, the Event MSRL head achieves 85.5\% AP50 and 43.3\% mAP in localization, compared with 46.0\% and 14.4\% for RGB. When classifying the same GT boxes, however, RGB exceeds Event by 4.1 Accuracy and 5.6 Macro-F1 points. The independent detectors reproduce this reversal, with Event leading in localization mAP by 27.8 points and RGB leading in Accuracy by 6.7 points and Macro-F1 by 10.2 points. Moreover, to rule out localization-coverage differences, we evaluate instances localized at $\text{IoU}\geq0.5$ by both RT-DETRs, where RGB still achieves 88.4\% Accuracy versus 82.9\% for Event. Further analyses across protocols, sizes, and densities are provided in the supplementary material.

\noindent\textbf{Cross-task recombination.}
We further keep the localization and objectness of one modality and replace its category evidence with that of the other. For the RT-DETRs, the category scores are exchanged between queries matched to the same GT object. As shown in Table~\ref{tab:fred-recombination}, Event localization with RGB categories performs best for both probes, and improves the RT-DETR results over Event categories by 12.2 AP50 and 6.1 mAP points. This further supports that the modality strengths differ across tasks.

\begin{table}[t]
\centering
\small
\setlength{\tabcolsep}{1.8pt}
\caption{Comparisons on the FRED canonical split.}
\label{tab:fred-canonical}
\begin{tabular*}{\columnwidth}{@{\extracolsep{\fill}}lclrr}
\toprule
Method & Venue & Input & AP50 & mAP \\
\midrule
Faster R-CNN~\tablecite{ren2015fasterrcnn} & NeurIPS'15 & RGB & 35.1 & 12.3 \\
RT-DETR~\tablecite{zhao2024rtdetr} & CVPR'24 & RGB & 34.1 & 11.8 \\
YOLOv11~\tablecite{khanam2024yolov11} & arXiv'24 & RGB & 35.2 & 13.4 \\
ER-DETR~\tablecite{magrini2024nerdd} & ECCVW'24 & RGB & 28.6 & 7.7 \\
YOLOv12~\tablecite{tian2025yolov12} & NeurIPS'25 & RGB & 35.2 & 13.5 \\
Mamba-YOLO~\tablecite{wang2025mambayolo} & AAAI'25 & RGB & 36.6 & 14.3 \\
\midrule
Faster R-CNN~\tablecite{ren2015fasterrcnn} & NeurIPS'15 & Event & 85.0 & 43.4 \\
RVT~\tablecite{gehrig2023rvt} & CVPR'23 & Event & 79.3 & 46.7 \\
SAST~\tablecite{peng2024sast} & CVPR'24 & Event & 77.9 & 45.0 \\
YOLOv11~\tablecite{khanam2024yolov11} & arXiv'24 & Event & \underline{87.7} & \underline{49.3} \\
RT-DETR~\tablecite{zhao2024rtdetr} & CVPR'24 & Event & 82.1 & 39.0 \\
ER-DETR~\tablecite{magrini2024nerdd} & ECCVW'24 & Event & 68.4 & 21.8 \\
SMamba~\tablecite{yang2025smamba} & AAAI'25 & Event & 81.7 & 47.7 \\
SparseVoxelDet~\tablecite{sadoun2026sparsevoxel} & arXiv'26 & Event & 84.3 & 40.2 \\
\midrule
CFT~\tablecite{fang2021cft} & arXiv'21 & RGB+Event & 85.3 & 41.4 \\
FPN-Fusion~\tablecite{tomy2022fpnfusion} & ICRA'22 & RGB+Event & 80.4 & 43.1 \\
RENet~\tablecite{zhou2023renet} & ICRA'23 & RGB+Event & 82.0 & 47.0 \\
SODFormer~\tablecite{li2023sodformer} & TPAMI'23 & RGB+Event & 82.4 & 48.1 \\
EOLO~\tablecite{cao2024eolo} & ICRA'24 & RGB+Event & 78.7 & 39.3 \\
ER-DETR~\tablecite{magrini2024nerdd} & ECCVW'24 & RGB+Event & 78.6 & 32.2 \\
ICAFusion~\tablecite{shen2024icafusion} & PR'24 & RGB+Event & 85.7 & 44.0 \\
PEPR~\tablecite{magrini2026pepr} & CVPRF'26 & RGB+Event & 38.2 & 11.9 \\
\midrule
AERODet & --- & RGB+Event & \textbf{88.5} & \textbf{49.7} \\
\bottomrule
\end{tabular*}
\end{table}
\noindent\textbf{Modality advantage gap.}
As shown in Table~\ref{tab:task-analysis}, the modality advantage gap on FRED is significantly larger than those on the RGB--IR datasets. On M3FD and VEDAI, the preferred modality switches between tasks, yet the margins are at most 1.9 mAP and 4.1 Macro-F1 points, compared with 27.8 and 10.2 points on FRED. On FLIR, IR leads in both tasks. Since Event data record brightness changes rather than rich appearance, the evidence gap between RGB and Event is much larger than that between RGB and IR, posing a greater challenge to fusion. This explains why task-symmetric fusion works well on RGB--IR data but cannot fully exploit the strengths of RGB--Event data.

\subsection{Comparison with State-of-the-Art}
\begin{table}[t]
\centering
\small
\setlength{\tabcolsep}{1.8pt}
\caption{Comparisons on the FRED challenging split.}
\label{tab:fred-challenging}
\begin{tabular*}{\columnwidth}{@{\extracolsep{\fill}}lclrr}
\toprule
Method & Venue & Input & AP50 & mAP \\
\midrule
Faster R-CNN~\tablecite{ren2015fasterrcnn} & NeurIPS'15 & RGB & 16.4 & 4.8 \\
YOLOv11~\tablecite{khanam2024yolov11} & arXiv'24 & RGB & 18.1 & 6.3 \\
RT-DETR~\tablecite{zhao2024rtdetr} & CVPR'24 & RGB & 21.1 & 7.1 \\
ER-DETR~\tablecite{magrini2024nerdd} & ECCVW'24 & RGB & 15.8 & 4.1 \\
\midrule
Faster R-CNN~\tablecite{ren2015fasterrcnn} & NeurIPS'15 & Event & 75.4 & 34.9 \\
YOLOv11~\tablecite{khanam2024yolov11} & arXiv'24 & Event & \underline{79.6} & \underline{41.6} \\
RT-DETR~\tablecite{zhao2024rtdetr} & CVPR'24 & Event & 76.9 & 35.1 \\
ER-DETR~\tablecite{magrini2024nerdd} & ECCVW'24 & Event & 62.2 & 19.8 \\
\midrule
CFT~\tablecite{fang2021cft} & arXiv'21 & RGB+Event & 76.9 & 33.5 \\
ER-DETR~\tablecite{magrini2024nerdd} & ECCVW'24 & RGB+Event & 75.7 & 27.9 \\
ICAFusion~\tablecite{shen2024icafusion} & PR'24 & RGB+Event & 76.0 & 33.7 \\
PEPR~\tablecite{magrini2026pepr} & CVPRF'26 & RGB+Event & 21.2 & 5.7 \\
\midrule
AERODet & --- & RGB+Event & \textbf{84.9} & \textbf{43.4} \\
\bottomrule
\end{tabular*}
\end{table}
\begin{table}[t]
\centering
\small
\setlength{\tabcolsep}{1.8pt}
\caption{Comparisons on the five-class FRED canonical split.}
\label{tab:fred-multiclass}
\begin{tabular*}{\columnwidth}{@{\extracolsep{\fill}}lclrr}
\toprule
Method & Venue & Input & AP50 & mAP \\
\midrule
YOLOv11~\tablecite{khanam2024yolov11} & arXiv'24 & RGB & 46.2 & 17.1 \\
RT-DETR~\tablecite{zhao2024rtdetr} & CVPR'24 & RGB & 47.7 & 17.0 \\
\midrule
YOLOv11~\tablecite{khanam2024yolov11} & arXiv'24 & Event & 74.2 & \underline{39.8} \\
RT-DETR~\tablecite{zhao2024rtdetr} & CVPR'24 & Event & 72.5 & 39.1 \\
\midrule
CFT~\tablecite{fang2021cft} & arXiv'21 & RGB+Event & \underline{74.4} & 35.1 \\
ICAFusion~\tablecite{shen2024icafusion} & PR'24 & RGB+Event & 73.7 & 36.3 \\
EOLO~\tablecite{cao2024eolo} & ICRA'24 & RGB+Event & 59.4 & 22.5 \\
M$^2$D-LIF~\tablecite{zhao2025m2dlif} & ICCV'25 & RGB+Event & 70.6 & 38.2 \\
SPFD~\tablecite{wang2026spfd} & CVPR'26 & RGB+Event & 67.4 & 28.5 \\
\midrule
AERODet & --- & RGB+Event & \textbf{82.1} & \textbf{42.8} \\
\bottomrule
\end{tabular*}
\end{table}
\begin{table}[t]
\centering
\small
\setlength{\tabcolsep}{1.8pt}
\caption{Comparisons on NeRDD.}
\label{tab:nerdd}
\begin{tabular*}{\columnwidth}{@{\extracolsep{\fill}}lclrr}
\toprule
Method & Venue & Input & AP50 & mAP \\
\midrule
YOLOv11~\tablecite{khanam2024yolov11} & arXiv'24 & RGB & 59.9 & 26.0 \\
YOLOv11~\tablecite{khanam2024yolov11} & arXiv'24 & Event & \underline{87.4} & \underline{51.0} \\
CFT~\tablecite{fang2021cft} & arXiv'21 & RGB+Event & 85.3 & 46.8 \\
ICAFusion~\tablecite{shen2024icafusion} & PR'24 & RGB+Event & 84.1 & 46.3 \\
EOLO~\tablecite{cao2024eolo} & ICRA'24 & RGB+Event & 71.6 & 33.6 \\
SPFD~\tablecite{wang2026spfd} & CVPR'26 & RGB+Event & 80.6 & 39.2 \\
\midrule
AERODet & --- & RGB+Event & \textbf{88.7} & \textbf{53.3} \\
\bottomrule
\end{tabular*}
\end{table}

\noindent\textbf{Evaluation on FRED.} As shown in Tables~\ref{tab:fred-canonical} and~\ref{tab:fred-challenging}, AERODet achieves 88.5\% AP50 and 49.7\% mAP on the canonical split and 84.9\% AP50 and 43.4\% mAP on the challenging split. On the canonical split, since Event data already provide strong localization cues for single-class detection under balanced conditions, existing RGB--Event methods fail to surpass the strongest Event-only detector, YOLOv11, whereas AERODet still outperforms it by 0.8 AP50 and 0.4 mAP points. On the challenging split, adverse lighting, rain, and insect distractors cause severe distribution shifts and modality degradation~\cite{magrini2025fred}. Under these conditions, the gains of AERODet increase to 5.3 AP50 and 1.8 mAP points over YOLOv11 and at least 8.0 AP50 and 9.7 mAP points over the RGB--Event methods, which could be ascribed to SURE adapting the modality contributions under modality-specific shifts.

\noindent\textbf{Evaluation under the five-class setting of FRED.} To further study the complementarity of RGB and Event, we evaluate drone-model detection with the five-class labels of FRED. As shown in Table~\ref{tab:fred-multiclass}, AERODet achieves 82.1\% AP50 and 42.8\% mAP, outperforming CFT by 7.7 AP50 points and M$^2$D-LIF by 4.6 mAP points. Notably, although the additional RGB input provides stronger category evidence as shown in Section~\ref{sec:task-analysis}, the RGB--Event baselines exceed the Event-only YOLOv11 by at most 0.2 AP50 points. Even the two strongest in AP50, CFT and ICAFusion, are 4.7 and 3.5 mAP points lower than YOLOv11, and M$^2$D-LIF, which enhances mono-modality learning, is 1.6 mAP points lower. As their fusion is shared by both tasks, it may be biased toward the Event evidence favored by localization, leaving the category evidence of RGB underexploited. In contrast, AERODet surpasses YOLOv11 by 7.9 AP50 points and 3.0 mAP points, far exceeding its margin of 0.8 AP50 points under the official single-class protocol. As the two settings share the same canonical split, this larger margin mainly reflects better category recognition, indicating that AERODet exploits the category evidence of RGB while retaining the localization advantage of Event.

\begin{figure*}[t]
\centering
\includegraphics[width=\textwidth]{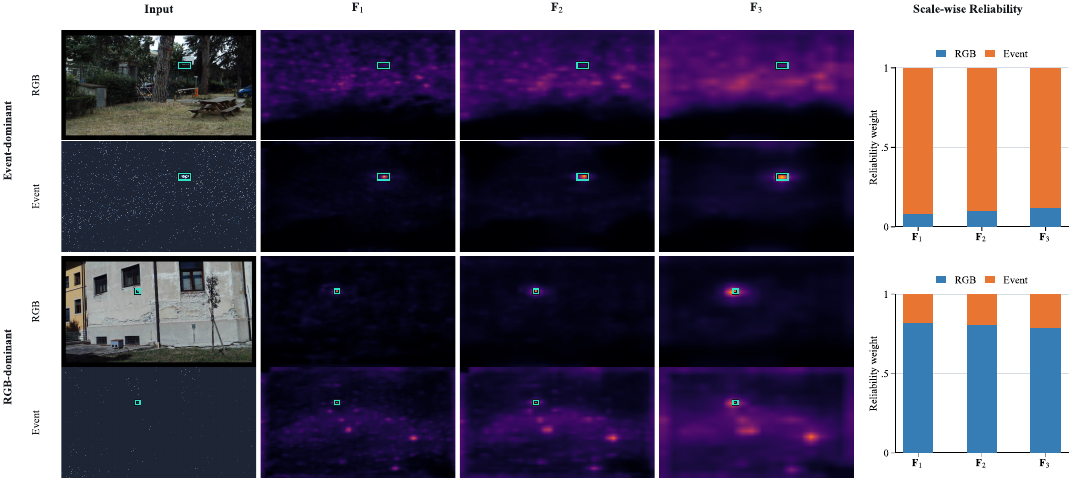}
\caption{Scale-wise reliability visualization of SURE. Cyan boxes denote ground truth.}
\label{fig:sure-responses}
\end{figure*}
\noindent\textbf{Evaluation on NeRDD.} As shown in Table~\ref{tab:nerdd}, AERODet achieves 88.7\% AP50 and 53.3\% mAP, outperforming the Event-only YOLOv11 by 1.3 AP50 points and 2.3 mAP points and CFT, the strongest RGB--Event method, by 3.4 AP50 points and 6.5 mAP points, which verifies its effectiveness across diverse scenes and backgrounds.
\begin{table}[t]
\centering\small
\setlength{\tabcolsep}{3pt}
\caption{Ablation study on SURE and DSOC. RGB, Event, and Joint denote the RoI features in RCI or the response in EOC, and $\checkmark$ denotes the default joint features and Event response.}
\label{tab:cumulative}
\begin{tabular*}{\columnwidth}{@{\extracolsep{\fill}}cccccrr}
\toprule
\multicolumn{2}{c}{SURE} & \multicolumn{2}{c}{DSOC} & & \multicolumn{2}{c}{Performance} \\
\cmidrule(lr){1-2}\cmidrule(lr){3-4}\cmidrule(lr){6-7}
MSRL & RGMC & RCI & EOC & Params (M) & AP50 & mAP \\
\midrule
$\times$ & $\times$ & $\times$ & $\times$ & 78.91 & 76.7 & 39.7 \\
$\checkmark$ & $\times$ & $\times$ & $\times$ & 79.80 & 78.6 & 40.8 \\
$\checkmark$ & $\checkmark$ & $\times$ & $\times$ & 79.80 & 79.4 & 41.1 \\
\midrule
$\checkmark$ & $\checkmark$ & RGB & $\times$ & 79.92 & 80.2 & 42.1 \\
$\checkmark$ & $\checkmark$ & RGB & $\checkmark$ & 79.92 & 80.8 & 42.2 \\
\midrule
$\checkmark$ & $\checkmark$ & Event & $\times$ & 79.92 & 79.9 & 41.8 \\
$\checkmark$ & $\checkmark$ & Event & $\checkmark$ & 79.92 & 80.5 & 41.9 \\
\midrule
$\checkmark$ & $\checkmark$ & $\checkmark$ & $\times$ & 79.99 & 81.5 & 42.7 \\
$\checkmark$ & $\checkmark$ & $\checkmark$ & RGB & 79.99 & 81.5 & 42.5 \\
$\checkmark$ & $\checkmark$ & $\checkmark$ & Joint & 79.99 & 82.0 & \textbf{42.8} \\
$\checkmark$ & $\checkmark$ & $\checkmark$ & $\checkmark$ & 79.99 & \textbf{82.1} & \textbf{42.8} \\
\bottomrule
\end{tabular*}
\end{table}

\subsection{Ablation Study}
We conduct ablations on the five-class FRED canonical split, as the challenging split has different class coverage between training and validation. We progressively introduce SURE and DSOC to examine the effect of each module.

\begin{table}[t]
\centering\small
\caption{Effect of RGMC under modality degradation on FRED.}
\label{tab:rgmc-degradation}
\begin{tabular*}{\columnwidth}{@{\extracolsep{\fill}}crr}
\toprule
RGMC & AP50 & mAP \\
\midrule
$\times$ & 48.1 & 17.5 \\
$\checkmark$ & \textbf{55.7} & \textbf{22.4} \\
\bottomrule
\end{tabular*}
\end{table}
\noindent\textbf{Effect of SURE.}
\label{sec:sure-analysis}
As shown in Table~\ref{tab:cumulative}, MSRL improves the baseline by 1.9 AP50 and 1.1 mAP points with 0.89M extra parameters. Since MSRL responses are not yet used at inference, this gain stems from independent supervision of each branch. RGMC further improves AP50 by 0.8 points without extra parameters. Although this gain on normal inputs is modest, RGMC mainly targets modality degradation. As shown in Table~\ref{tab:rgmc-degradation}, under modality degradation, turning off RGMC drops AP50 from 55.7\% to 48.1\% and mAP from 22.4\% to 17.5\%, as detailed in the supplementary material. Figure~\ref{fig:sure-responses} shows that SURE favors Event when its responses concentrate on the target while RGB responses spread over the textured background, and shifts toward RGB when the Event frame is nearly empty. More examples are provided in the supplementary material.

\noindent\textbf{Effect of DSOC.}
As shown in Table~\ref{tab:cumulative}, injecting RGB RoI features in RCI raises AP50 from 79.4\% to 80.2\% and mAP from 41.1\% to 42.1\%. Since RCI complements the decoupled query rather than classifying RoI features standalone, this confirms that the shared query loses classification-favorable cues, especially RGB category evidence. The joint token of RCI further raises AP50 to 81.5\% and mAP to 42.7\% with 0.07M extra parameters, outperforming single-modality variants and the parameter-matched control in the supplementary material. Moreover, EOC with the Event RoI response improves AP50 by 0.6 points, while the RGB and averaged responses only reach 81.5\% and 82.0\%. Since EOC only rescores candidates without changing their boxes, this gain shows that the Event RoI response ranks foreground above background candidates, consistent with its role as a foreground indicator. Overall, SURE and DSOC improve the baseline by 5.4 AP50 and 3.1 mAP points with only 1.08M extra parameters (1.4\%).
\section{Conclusion}
In this paper, we rethink task-symmetric fusion in RGB--Event moving object detection and reveal a task-dependent modality advantage reversal. Specifically, Event provides stronger localization, whereas RGB offers stronger category evidence for classification. To exploit the strengths of each modality, we propose AERODet, which allows localization and classification to adopt different fusion strategies. It uses SURE to calibrate localization weights via supervised unimodal responses, and introduces DSOC to decouple classification for semantic and objectness calibration. Experiments on FRED and NeRDD validate the effectiveness of our design. In the future, we hope to extend this asymmetric paradigm to other multimodal settings with large sensor gaps, such as RGB--LiDAR, as well as broader vision tasks including instance segmentation and multi-object tracking.
\section*{LLM Usage Statement}
LLM tools were used in four ways, and only to edit or refine content produced by the authors. First, for writing, we used OpenAI Codex and Anthropic Claude in Cursor to edit the language and restructure the text drafted by the authors, and we used OpenAI ChatGPT to obtain reviewer-style feedback on the manuscript. Second, for related work, these tools assisted in checking the citations and their bibliographic metadata. Third, for code, experiments, and figures, these tools assisted in refining the authors' analysis scripts, result tables, LaTeX source, and plotting code. Fourth, for the research ideas, claims, and conclusions, these tools were only used to refine the presentation of the authors' own formulations. No LLM or AI tool was used to create or annotate data. The authors take full responsibility for all submitted content, regardless of any LLM or AI-tool use.
{
    \small
    \fontsize{9}{10.1}\selectfont
    \setlength{\bibsep}{0.2pt plus 0.2pt}
    \bibliographystyle{ieeenat_fullname}
    \bibliography{main}
}

\clearpage
\makeatletter
\setlength{\@fptop}{0pt}
\setlength{\@fpsep}{16pt}
\setlength{\@fpbot}{0pt plus 1fil}
\setlength{\@dblfptop}{0pt}
\setlength{\@dblfpsep}{16pt}
\setlength{\@dblfpbot}{0pt plus 1fil}
\makeatother
\maketitlesupplementary
\setcounter{section}{0}
\setcounter{subsection}{0}
\setcounter{subsubsection}{0}
\setcounter{table}{0}
\setcounter{figure}{0}
\setcounter{equation}{0}
\providecommand{\theHsection}{}
\renewcommand{\theHsection}{supp.\arabic{section}}
\providecommand{\theHsubsection}{}
\renewcommand{\theHsubsection}{supp.\arabic{section}.\arabic{subsection}}
\providecommand{\theHsubsubsection}{}
\renewcommand{\theHsubsubsection}{supp.\arabic{section}.\arabic{subsection}.\arabic{subsubsection}}
\providecommand{\theHtable}{}
\renewcommand{\theHtable}{supp.\arabic{table}}
\providecommand{\theHfigure}{}
\renewcommand{\theHfigure}{supp.\arabic{figure}}
\providecommand{\theHequation}{}
\renewcommand{\theHequation}{supp.\arabic{equation}}
\renewcommand{\thesection}{S\arabic{section}}
\renewcommand{\thetable}{S\arabic{table}}
\renewcommand{\thefigure}{S\arabic{figure}}
\renewcommand{\theequation}{S\arabic{equation}}

This supplementary material provides additional technical details and empirical analyses supporting the main paper. Section~\ref{sec:s-method} details the architecture, training configuration, and attention-bias derivation of RGMC. Section~\ref{sec:s-analysis} extends our task-dependent modality analysis on FRED and external RGB--Infrared benchmarks. Section~\ref{sec:s-comparisons} reports additional baseline comparisons on FRED and evaluates generalization to multi-class road-object detection on PEOD. Section~\ref{sec:s-ablations} presents further ablations and hyperparameter sensitivity analyses for SURE and DSOC. Finally, Section~\ref{sec:s-qual} provides additional qualitative visualizations of modality reliability.

\section{Method and Implementation Details}
\label{sec:s-method}

\subsection{Implementation Settings}
The RGB and Event branches produce three encoder feature maps $\{\mathbf F_l^m\}_{l=1}^{3}$ for each modality $m\in\{rgb,evt\}$, ordered from highest to lowest spatial resolution. The six maps form the decoder memory~\cite{zhao2024rtdetr,guo2024damsdet}. Table~\ref{tab:s-architecture} summarizes the architecture.
\begin{table}[!htb]
\centering\small
\setlength{\tabcolsep}{3pt}
\caption{Architecture settings of AERODet.}
\label{tab:s-architecture}
\begin{tabular*}{\columnwidth}{@{\extracolsep{\fill}}>{\raggedright\arraybackslash}p{0.31\columnwidth}>{\raggedright\arraybackslash}p{0.63\columnwidth}}
\toprule
Component & Setting \\
\midrule
Feature branches & Separate ResNet-50-vd and HybridEncoder parameters for RGB and Event \\
Feature maps & $\mathbf F_1^m$, $\mathbf F_2^m$, $\mathbf F_3^m$ with 256 channels each and strides 8, 16, and 32 \\
Decoder memory & Six modality--scale maps with separate projections \\
Queries & 300 matching queries, denoising budget 100 (generated count follows the batch GT number) \\
Decoder & 6 layers, 8 heads, 4 points per modality--scale combination and head \\
MSRL heads & One heatmap, offset, and size head per modality, shared across the three scales \\
Task adapters & Two $256\to64\to256$ residual adapters \\
Regional features & $3\times3$ RoIAlign on $\mathbf F_1^{rgb}$ and $\mathbf F_1^{evt}$, followed by average pooling \\
Joint projection & Two 256-dimensional tokens concatenated, $512\to256$ \\
Category heads & Single-foreground main head, $C$-category MSRL and RCI heads \\
\bottomrule
\end{tabular*}
\end{table}

For numerical stability, we use $\epsilon=10^{-6}$ to protect denominators and logarithm arguments where needed, including in response normalization, entropy, and the reliability ratio. Equations omit these safeguards for clarity. In EOC, we floor the RoI responses at $10^{-4}$ before computing their mean and clip the final detection scores to $[10^{-6},1-10^{-6}]$ before converting them to logits.

\subsection{MSRL Heads and Supervision}
Each branch of the MSRL head, $\Psi_{hm}$, $\Psi_o$, or $\Psi_s$, uses a $3\times3$ convolution with 64 output channels, ReLU, and a $1\times1$ output convolution~\cite{zhou2019objects}. The branches predict $\hat{\mathbf Y}$, $\hat{\mathbf O}$, and $\hat{\mathbf S}$ with $C$, 2, and 4 channels, respectively. The heatmap bias is initialized to $-2.19$.

At each scale, we use the same targets for the aligned RGB and Event frames. We place a Gaussian peak at the projected object center, using a minimum overlap of 0.7 to determine its radius~\cite{zhou2019objects}. The offset gives the displacement from the grid location to the continuous center, and the size target gives the distances from that center to the four box boundaries. The three branches supervise their modality's encoder and backbone. RGMC and EOC read the detached objectness map $\mathbf H_l^m$, obtained by taking the maximum of $\hat{\mathbf Y}$ over the category channels.

\subsection{Response-Guided Modality Calibration}
For a fixed query and attention head in a decoder layer, let $W_{l,s}^m$ denote the deformable cross-attention logit~\cite{zhu2021deformabledetr} for sampling point $s$ on $\mathbf F_l^m$. The uncalibrated weights $A_{l,s}^m$ share a softmax over the $2\times3\times4=24$ sampling slots,
\begin{equation}
A_{l,s}^m=\frac{\exp(W_{l,s}^m)}{\sum_{m',l',s'}\exp(W_{l',s'}^{m'})}.
\end{equation}
The main paper defines the pre-normalization multiplier $\omega_l^m=(\alpha_l^m/\alpha_l^{\bar m})^{1/2}$, where $\bar m$ denotes the other modality. Since $\omega_l^m\exp(W_{l,s}^m)=\exp(W_{l,s}^m+\log\omega_l^m)$, the multiplier becomes an additive bias before softmax. The two multipliers are reciprocal, so their log biases have opposite signs,
\begin{equation}
\begin{gathered}
b_l=\frac12\log\frac{\alpha_l^{evt}}{\alpha_l^{rgb}},\\
\tilde{W}_{l,s}^{evt}=W_{l,s}^{evt}+b_l,\quad
\tilde{W}_{l,s}^{rgb}=W_{l,s}^{rgb}-b_l.
\end{gathered}
\end{equation}
Applying the joint softmax to $\tilde{W}_{l,s}^m$ gives the calibrated weights $\tilde{A}_{l,s}^m$. The shared normalizer cancels in the Event-to-RGB ratio,
\begin{equation}
\frac{\tilde{A}_{l,s}^{evt}}{\tilde{A}_{l,s}^{rgb}}
=\frac{A_{l,s}^{evt}}{A_{l,s}^{rgb}}e^{2b_l}
=\frac{A_{l,s}^{evt}}{A_{l,s}^{rgb}}\frac{\alpha_l^{evt}}{\alpha_l^{rgb}}.
\end{equation}
Within each $\mathbf F_l^m$, all points share the same bias. Their relative weights and sampling offsets therefore remain unchanged. Each head aggregates its sampled features with $\tilde{A}_{l,s}^m$, and the concatenated head outputs are projected to update the query.

With $\eta=0.01$ and $640\times640$ inputs, RGMC retains 64, 16, and 4 positions on the response maps $\mathbf H_1^m$, $\mathbf H_2^m$, and $\mathbf H_3^m$, respectively. The same calibration is applied to decoder cross-attention in training and inference, with detached responses and reliabilities. Encoder query selection uses the original confidence scores.

\subsection{Disentangled Semantic and Objectness Calibration}
At the final decoder layer, $\operatorname{Adapter}_{loc}$ and $\operatorname{Adapter}_{obj}$ are bottleneck MLPs ($256\to64\to256$ with ReLU). Their output projections are initialized to zero, so $\mathbf q^{loc}$ and $\mathbf q^{obj}$ initially reduce to the layer-normalized decoder query. Their heads then predict the box residual and the single-foreground objectness logit, respectively.

RCI extracts aligned $3\times3$ RoI features from $\mathbf F_1^{rgb}$ and $\mathbf F_1^{evt}$ using a sampling ratio of 2~\cite{he2017maskrcnn}. The pooled tokens are projected by $\operatorname{Proj}_{joint}$ and injected into the detached $\mathbf q^{obj}$ to form $\mathbf q^{cls}$. The Varifocal targets~\cite{zhang2021varifocal} inherit the Hungarian assignments and matched-box IoUs from the last decoder layer. Unmatched queries contribute as negatives, and denoising queries are excluded from $\mathcal L_{\mathrm{rci}}$. This loss updates only the joint projection, the layer normalization of $\mathbf q^{cls}$, and the category head.

EOC computes $\bar r$ over the 300 query responses of each image. At inference, we retain the 100 query--category pairs with the highest $\hat s_{j,c}^{det}$. The RGB and Event variants in the main-paper ablation use $\mathbf H_1^{rgb}$ and $\mathbf H_1^{evt}$, respectively. The Joint variant averages the two log-ratio terms after normalizing each modality's response by its own query mean. All three variants use $\rho=0.5$.

\subsection{Training and Evaluation}
Table~\ref{tab:s-training} summarizes the training hyperparameters of AERODet. Following the standard COCO evaluation protocol, we report AP50 at IoU 0.50 and mAP averaged over IoU thresholds from 0.50 to 0.95 with a step size of 0.05. For class-agnostic localization evaluation, multi-class predictions are collapsed into a single foreground class at test time.
\begin{table}[t]
\centering\small
\setlength{\tabcolsep}{3pt}
\caption{Training hyperparameters.}
\label{tab:s-training}
\begin{tabular*}{\columnwidth}{@{\extracolsep{\fill}}>{\raggedright\arraybackslash}p{0.31\columnwidth}>{\raggedright\arraybackslash}p{0.63\columnwidth}}
\toprule
Item & Setting \\
\midrule
Schedule & 5 epochs \\
Initialization & Shared COCO-pretrained weights \\
Optimizer & AdamW, learning rate $10^{-4}$, weight decay $10^{-4}$ \\
Backbone rate & $10^{-5}$ for ResNet stages \\
Batch size & 4 image pairs \\
Warmup & 2,000 steps from a factor of 0.001 \\
LR schedule & Constant after warmup \\
Gradient clipping & Global norm 0.1 \\
Moving average & Exponential moving average, decay 0.9999 \\
Train resolution & Multiscale square inputs, 480--800 pixels \\
Test resolution & $640\times640$ \\
Augmentation & Paired geometric transforms, distortion, expansion, crop, and horizontal flip \\
Matcher costs & Objectness 2, $L_1$ 5, generalized IoU 2 \\
Detection losses & Objectness 1, $L_1$ 5, generalized IoU 2 \\
Denoising & Budget 100, label noise 0.5, box noise 1.0 \\
Module losses & $\lambda_{\mathrm{msrl}}=1$, $\lambda_{\mathrm{rci}}=0.25$ \\
\bottomrule
\end{tabular*}
\end{table}
\FloatBarrier

\section{Extended Task-Dependent Modality Analysis}
\label{sec:s-analysis}

\subsection{FRED Modality Analysis}
\paragraph{Readout Protocols.}
\noindent\textbf{MSRL heads.} On FRED~\cite{magrini2025fred}, we probe the RGB and Event MSRL heads of a task-symmetric baseline trained with MSRL alone (without RGMC and DSOC). Each head decodes its top 100 predictions with score threshold 0.001 and NMS threshold 0.6. Class-agnostic localization ignores predicted categories. Fixed-box classification reads the peak response of each category within the same 134,486 ground-truth boxes and selects the highest-scoring class, whereas predicted-box classification evaluates responses inside predictions matched to ground truth at IoU $\geq0.5$ (90,910 RGB and 126,118 Event instances). Cross-task recombination pairs the boxes and foreground scores of one head with the five category responses of either head on those candidates.

\noindent\textbf{Independent detectors.} We train RGB-only and Event-only RT-DETRs under the same five-class protocol. In class-agnostic localization, each query uses its maximum category score as objectness. For matched-query classification, we assign the 300 decoder queries to ground-truth objects using Hungarian matching based only on IoU. The all-object readout includes low-IoU matches. In cross-task recombination, we keep the box and foreground confidence from one detector and take the category scores from the other detector's query matched to the same object. The resulting AP is computed on the ground-truth-matched queries.

\paragraph{Relative Reliability of Auxiliary Responses.}
We test whether the first-scale reliability predicts which MSRL head localizes better. For each modality, we compute the IoU between the top-scoring box from $\mathbf F_1^m$ and ground truth. Over the 116,801 FRED images with unequal IoUs, we rank RGB wins by $\Delta\alpha=\alpha_1^{rgb}-\alpha_1^{evt}$. As shown in Table~\ref{tab:s-reliability-ranking}, SURE reaches 0.705 AUROC and 0.347 precision--recall AP. Its Spearman correlation with the signed IoU gap is 0.486. AUROC rises to 0.812 on the 82,319 images with an absolute IoU gap of at least 0.10 and to 0.907 on the 56,757 images with a gap of at least 0.20. Selecting the auxiliary box from the modality with higher reliability yields a mean top-1 IoU of 0.724, compared with 0.719 when always selecting Event and 0.709 when selecting by peak-response difference.
\begin{table}[t]
\centering\small
\setlength{\tabcolsep}{3pt}
\caption{Modality reliability ranking at scale $l=1$ on FRED. PR AP treats RGB wins as positives.}
\label{tab:s-reliability-ranking}
\begin{tabular*}{\columnwidth}{@{\extracolsep{\fill}}lrr}
\toprule
Ranking score & AUROC & PR AP \\
\midrule
Log peak ratio & 0.678 & 0.292 \\
SURE & \textbf{0.705} & \textbf{0.347} \\
\bottomrule
\end{tabular*}
\end{table}

\paragraph{Classification Readouts.}
Table~\ref{tab:s-readouts} extends the independent-detector comparison across matched-query IoU thresholds, decoded predictions, and the common subset of 100,064 objects localized at IoU $\geq0.5$ by both modalities. RGB outperforms Event under every readout, even on the common subset where localization coverage is identical. The MSRL heads confirm the same trend on matched predictions, where RGB reaches 90.3\% Accuracy and 90.1\% Macro-F1 versus 84.6\% and 82.4\% for Event.
\begin{table}[t]
\centering\small
\setlength{\tabcolsep}{2.5pt}
\caption{Classification of independent detectors on FRED across readout protocols and IoU thresholds.}
\label{tab:s-readouts}
\begin{tabular*}{\columnwidth}{@{\extracolsep{\fill}}lllrrr}
\toprule
Readout & Min. IoU & Modality & Objects & Accuracy & Macro-F1 \\
\midrule
\multirow{6}{*}{Matched} & \multirow{2}{*}{All} & RGB & 134,486 & \textbf{87.2} & \textbf{86.4} \\
 & & Event & 134,486 & 80.5 & 76.2 \\
  & \multirow{2}{*}{0.30} & RGB & 122,367 & \textbf{87.6} & \textbf{86.7} \\
 & & Event & 132,777 & 80.8 & 76.5 \\
  & \multirow{2}{*}{0.50} & RGB & 101,913 & \textbf{87.6} & \textbf{86.5} \\
 & & Event & 128,336 & 81.4 & 77.3 \\
\midrule
\multirow{4}{*}{Decoded} & \multirow{2}{*}{0.50} & RGB & 97,507 & \textbf{88.2} & \textbf{87.1} \\
 & & Event & 127,387 & 82.7 & 78.7 \\
  & \multirow{2}{*}{0.75} & RGB & 36,305 & \textbf{90.4} & \textbf{88.2} \\
 & & Event & 92,254 & 84.1 & 79.3 \\
\midrule
\multirow{2}{*}{Common} & \multirow{2}{*}{0.50} & RGB & 100,064 & \textbf{88.4} & --- \\
 & & Event & 100,064 & 82.9 & --- \\
\bottomrule
\end{tabular*}
\end{table}

\paragraph{Grouped Results.}
Table~\ref{tab:s-groups} breaks down matched IoU and common-object classification accuracy by drone model, object scale, and Event density. Event has the higher matched IoU in every group. RGB has the higher accuracy in every group except Tello EDU, where both modalities exceed 96\%.
\begin{table}[t]
\centering\small
\setlength{\tabcolsep}{2.5pt}
\caption{Grouped FRED diagnostics by drone category, object size, and Event density. IoU is reported in $[0,1]$.}
\label{tab:s-groups}
\begin{tabular*}{\columnwidth}{@{\extracolsep{\fill}}llrrrrr}
\toprule
 & & & \multicolumn{2}{c}{Matched IoU} & \multicolumn{2}{c}{Accuracy} \\
\cmidrule(lr){4-5}\cmidrule(lr){6-7}
Factor & Group & Objects & RGB & Event & RGB & Event \\
\midrule
\multirow{5}{*}{Category} & Tello EDU & 55,314 & 0.652 & \textbf{0.794} & 96.5 & \textbf{98.5} \\
 & CineApe20 & 38,873 & 0.558 & \textbf{0.774} & \textbf{80.8} & 68.5 \\
 & Air75 & 18,531 & 0.536 & \textbf{0.749} & \textbf{78.3} & 68.0 \\
 & Mini 2 & 9,752 & 0.666 & \textbf{0.807} & \textbf{91.3} & 77.7 \\
 & Mini 3 & 12,016 & 0.652 & \textbf{0.719} & \textbf{80.7} & 72.6 \\
\midrule
\multirow{3}{*}{Size} & Small & 44,829 & 0.516 & \textbf{0.738} & \textbf{92.9} & 85.0 \\
 & Medium & 44,369 & 0.652 & \textbf{0.793} & \textbf{89.6} & 83.6 \\
 & Large & 45,288 & 0.662 & \textbf{0.798} & \textbf{84.0} & 80.7 \\
\midrule
\multirow{3}{*}{Density} & Low & 53,900 & 0.595 & \textbf{0.760} & \textbf{79.4} & 70.7 \\
 & Mid & 45,377 & 0.612 & \textbf{0.775} & \textbf{92.3} & 87.4 \\
 & High & 35,209 & 0.631 & \textbf{0.803} & \textbf{96.2} & 94.9 \\
\bottomrule
\end{tabular*}
\end{table}

\begin{figure*}[t]
\centering
\includegraphics[width=\textwidth]{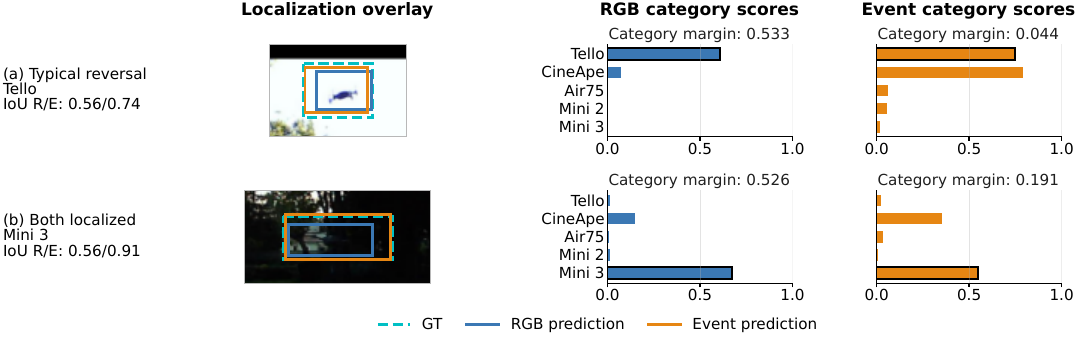}
\caption{Probe cases and regional category responses on FRED. Dashed cyan boxes denote ground truth, and blue and orange boxes denote RGB and Event predictions. The category margin is the gap between the top two class scores.}
\label{fig:s-probe-cases}
\end{figure*}

\paragraph{Probe Cases.}
Figure~\ref{fig:s-probe-cases} shows two FRED cases in which both modalities classify the same regions. In the first case, Event yields a tighter box but assigns a slightly higher score to an incorrect category. RGB identifies the correct category with a wider margin. In the second case, both modalities localize the drone, and RGB again assigns the correct category a wider margin.

\subsection{External RGB--IR Dataset Analysis}
\paragraph{Datasets and Independent Detectors.}
We use 2,940 training and 1,260 evaluation pairs for M3FD~\cite{liu2022tar}, 4,129 and 1,013 for FLIR~\cite{zhang2020flir}, and 1,089 and 121 for the first fold of VEDAI~\cite{razakarivony2016vedai}. The corresponding numbers of categories are six, three, and eleven. FLIR uses the aligned split. M3FD and FLIR pair RGB with Thermal, and VEDAI pairs RGB with near-infrared (NIR).

For each modality, we train an independent RT-DETR v1 with a ResNet-50-vd backbone from COCO initialization. We train for 40 epochs with a batch size of 4 and initial learning rates of $10^{-5}$ for the backbone and $10^{-4}$ elsewhere, decaying both at epoch 30. We evaluate the checkpoint with the highest validation mAP. Following Table~1 of the main paper, IR denotes Thermal on M3FD and FLIR, and NIR on VEDAI.

\paragraph{Matched-Query Classification.}
We apply the FRED matched-query protocol to each external dataset. Table~\ref{tab:s-ext-query} includes the common subset, where both detectors match the same ground-truth object at IoU $\geq0.5$. Thermal leads on M3FD and RGB leads on VEDAI on both subsets. On FLIR, the two modalities reach virtually identical Macro-F1 on the common subset.
\begin{table}[t]
\centering\small
\setlength{\tabcolsep}{2pt}
\caption{Matched-query classification on the external RGB--IR datasets.}
\label{tab:s-ext-query}
\begin{tabular*}{\columnwidth}{@{\extracolsep{\fill}}lllrrr}
\toprule
Dataset & Objects & Modality & Count & Accuracy & Macro-F1 \\
\midrule
\multirow{4}{*}{M3FD} & \multirow{2}{*}{All} & RGB & 10,515 & 94.5 & 85.1 \\
 & & Thermal & 10,515 & \textbf{96.8} & \textbf{89.2} \\
  & \multirow{2}{*}{Common} & RGB & 9,764 & 94.9 & 85.9 \\
 & & Thermal & 9,764 & \textbf{97.5} & \textbf{90.9} \\
\midrule
\multirow{4}{*}{FLIR} & \multirow{2}{*}{All} & RGB & 8,591 & 97.8 & 96.0 \\
 & & Thermal & 8,591 & \textbf{98.7} & \textbf{96.5} \\
  & \multirow{2}{*}{Common} & RGB & 8,267 & 98.2 & \textbf{96.6} \\
 & & Thermal & 8,267 & \textbf{98.8} & \textbf{96.6} \\
\midrule
\multirow{4}{*}{VEDAI} & \multirow{2}{*}{All} & RGB & 366 & \textbf{74.6} & \textbf{62.3} \\
 & & NIR & 366 & 72.7 & 60.3 \\
  & \multirow{2}{*}{Common} & RGB & 363 & \textbf{74.7} & \textbf{62.4} \\
 & & NIR & 363 & 72.7 & 60.5 \\
\bottomrule
\end{tabular*}
\end{table}

\paragraph{Cross-Task Recombination.}
We apply the same cross-task recombination protocol to the independent RGB and Infrared detectors. As shown in Table~\ref{tab:s-ext-recomb}, the strongest combination uses Thermal for both tasks on M3FD and FLIR, and RGB for both tasks on VEDAI. This provides a counterpart to the FRED probe, where the combination of Event localization and RGB category evidence yields the highest diagnostic AP.
\begin{table}[t]
\centering\small
\setlength{\tabcolsep}{4pt}
\caption{Cross-task recombination on the external datasets (diagnostic AP on ground-truth-matched queries).}
\label{tab:s-ext-recomb}
\begin{tabular*}{\columnwidth}{@{\extracolsep{\fill}}lllrr}
\toprule
Dataset & Localization & Category & AP50 & mAP \\
\midrule
\multirow{4}{*}{M3FD} & RGB & RGB & 74.0 & 49.3 \\
 & RGB & Thermal & 64.8 & 43.7 \\
 & Thermal & RGB & 59.0 & 38.8 \\
 & Thermal & Thermal & \textbf{78.9} & \textbf{50.2} \\
\midrule
\multirow{4}{*}{FLIR} & RGB & RGB & 88.3 & 44.4 \\
 & RGB & Thermal & 78.5 & 40.4 \\
 & Thermal & RGB & 84.2 & 48.9 \\
 & Thermal & Thermal & \textbf{90.7} & \textbf{52.0} \\
\midrule
\multirow{4}{*}{VEDAI} & RGB & RGB & \textbf{65.3} & \textbf{43.6} \\
 & RGB & NIR & 45.2 & 28.7 \\
 & NIR & RGB & 31.2 & 20.6 \\
 & NIR & NIR & 42.0 & 28.0 \\
\bottomrule
\end{tabular*}
\end{table}

\paragraph{Classification within Fixed Regions.}
To isolate regional category evidence, we apply aligned $3\times3$ RoIAlign to the highest-resolution encoder map of each frozen independent detector, using the same ground-truth boxes for both modalities. We average-pool the RoIs and train classifiers with one 128-unit hidden layer and GELU for 20 epochs, using AdamW with a learning rate of $10^{-3}$ and a batch size of 256. The $\times2$ controls duplicate a single modality's token to match the input width and parameter count of the Joint classifier.

As shown in Table~\ref{tab:s-ext-regions}, within identical ground-truth regions, the single-modality classifiers differ by at most 1.1 points in Accuracy and 2.0 points in Macro-F1 across the three RGB--IR benchmarks. The gaps are substantially smaller than those between RGB and Event on FRED. The Joint regional representation improves Macro-F1 over the stronger parameter-matched single-modality control on every dataset, supporting the use of joint regional semantic features.
\begin{table}[t]
\centering\small
\setlength{\tabcolsep}{2pt}
\caption{Fixed-region classification on the external datasets. $\times2$ duplicates one modality's token to match the Joint parameter count.}
\label{tab:s-ext-regions}
\begin{tabular*}{\columnwidth}{@{\extracolsep{\fill}}llrrr}
\toprule
Dataset & Regional features & Params & Accuracy & Macro-F1 \\
\midrule
\multirow{5}{*}{M3FD} & RGB & 33,670 & 97.8 & 94.3 \\
 & Thermal & 33,670 & 97.5 & 92.3 \\
 & RGB$\times$2 & 66,438 & 97.9 & 94.9 \\
 & Thermal$\times$2 & 66,438 & 97.6 & 92.7 \\
 & Joint & 66,438 & \textbf{98.3} & \textbf{95.0} \\
\midrule
\multirow{5}{*}{FLIR} & RGB & 33,283 & 98.2 & 97.1 \\
 & Thermal & 33,283 & 98.9 & 97.5 \\
 & RGB$\times$2 & 66,051 & 98.2 & 97.1 \\
 & Thermal$\times$2 & 66,051 & 98.9 & 97.4 \\
 & Joint & 66,051 & \textbf{99.2} & \textbf{98.0} \\
\midrule
\multirow{5}{*}{VEDAI} & RGB & 34,315 & 86.3 & 71.8 \\
 & NIR & 34,315 & 85.2 & 70.1 \\
 & RGB$\times$2 & 67,083 & \textbf{86.9} & 71.6 \\
 & NIR$\times$2 & 67,083 & 85.8 & 70.5 \\
 & Joint & 67,083 & \textbf{86.9} & \textbf{71.9} \\
\bottomrule
\end{tabular*}
\end{table}

\section{Additional Comparisons and Cross-Scene Generalization}
\label{sec:s-comparisons}

\subsection{Additional Baselines on FRED}
We further compare AERODet with FlexEvent~\cite{lu2025flexevent}, MMFD~\cite{guo2026mmfd}, and InfraNet~\cite{feng2026infranet} on the five-class FRED canonical split. We adapt InfraNet with RGB guidance during training and Event-only inference. We also construct a two-stage asymmetric baseline that localizes objects with Event features and classifies the resulting boxes with RGB RoI features. As shown in Table~\ref{tab:s-additional-fred}, AERODet leads all listed methods on both metrics and exceeds the two-stage baseline by 7.4 AP50 points and 4.7 mAP points, supporting end-to-end asymmetric fusion.

\begin{table}[!htb]
\centering\small
\setlength{\tabcolsep}{3pt}
\caption{Additional baselines on five-class FRED.}
\label{tab:s-additional-fred}
\begin{tabular*}{\columnwidth}{@{\extracolsep{\fill}}lcrr}
\toprule
Method & Venue & AP50 & mAP \\
\midrule
FlexEvent~\cite{lu2025flexevent} & NeurIPS'25 & 69.9 & 36.1 \\
MMFD~\cite{guo2026mmfd} & TMM'26 & 74.2 & 39.3 \\
InfraNet~\cite{feng2026infranet} & ECCV'26 & 68.8 & 35.9 \\
Event localization + RGB RoI & --- & 74.7 & 38.1 \\
AERODet & --- & \textbf{82.1} & \textbf{42.8} \\
\bottomrule
\end{tabular*}
\end{table}

\subsection{Generalization to Road-Object Detection on PEOD}
To evaluate generalization beyond aerial drone detection, we further test AERODet on PEOD~\cite{cui2026peod}, a pixel-aligned RGB--Event benchmark for six-category road-object detection. Following the official split, we train on 59,628 pairs and evaluate on all 11,847 test pairs, including 5,444 normal and 6,403 challenging pairs. As shown in Table~\ref{tab:s-peod}, AERODet achieves 63.7\% AP50 and 41.7\% mAP, improving on SPFD~\cite{wang2026spfd} by 17.3 AP50 points and 15.5 mAP points and exceeding all listed unimodal and RGB--Event baselines.

We also probe regional category evidence on PEOD by reading category scores from both MSRL heads within identical Event-localized boxes. RGB reaches 83.5\% top-1 accuracy, compared with 81.3\% for Event. The RGB advantage in regional classification thus extends from aerial targets to road objects.

\begin{table}[t]
\centering\small
\setlength{\tabcolsep}{1.7pt}
\caption{Six-class road-object detection on PEOD. Published results are taken from Table~2 of~\cite{cui2026peod}. $\dagger$ denotes our evaluation of SPFD under the same protocol.}
\label{tab:s-peod}
\begin{tabular*}{\columnwidth}{@{\extracolsep{\fill}}lclrr}
\toprule
Method & Venue & Input & AP50 & mAP \\
\midrule
RVT & CVPR'23 & Event & 38.4 & 20.1 \\
SAST & CVPR'24 & Event & 37.8 & 18.1 \\
SMamba & AAAI'25 & Event & 43.8 & 22.9 \\
RT-DETR & CVPR'24 & RGB & 38.5 & 21.8 \\
\midrule
FPN-Fusion & ICRA'22 & RGB+Event & 29.9 & 14.6 \\
RENet & ICRA'23 & RGB+Event & 41.6 & 24.5 \\
SODFormer & TPAMI'23 & RGB+Event & 36.3 & 17.8 \\
EOLO & ICRA'24 & RGB+Event & 45.8 & 26.1 \\
SFNet & TITS'24 & RGB+Event & 37.9 & 19.5 \\
CAFR & ECCV'24 & RGB+Event & 39.6 & 21.4 \\
\midrule
SPFD$^\dagger$ & CVPR'26 & RGB+Event & 46.4 & 26.2 \\
AERODet & --- & RGB+Event & \textbf{63.7} & \textbf{41.7} \\
\bottomrule
\end{tabular*}
\end{table}

\begin{figure*}[t]
\centering
\setlength{\unitlength}{\dimexpr\textwidth/495\relax}
\begin{picture}(495,136.45)
  \put(80,0){\includegraphics[trim=80bp 0bp 0bp 0bp,clip,width=414.712\unitlength]{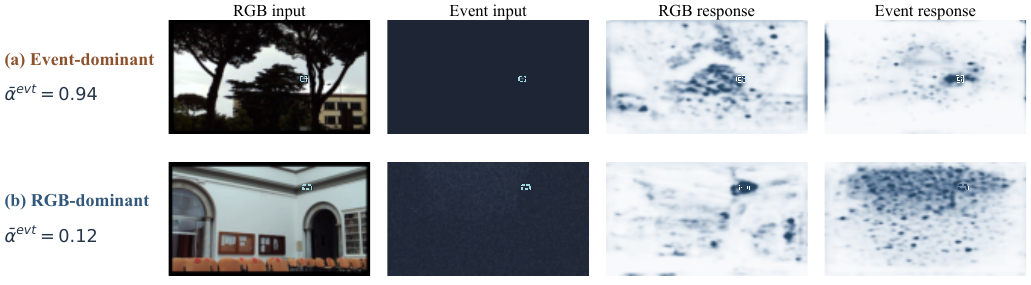}}
  \put(2.16,105.13){{\small \bfseries\color[HTML]{8C522E}(a) Event-dominant}}
  \put(2.16,88.28){{\small \color[HTML]{263649}$\tilde{\alpha}^{evt}=0.94$}}
  \put(2.16,37.09){{\small \bfseries\color[HTML]{31577A}(b) RGB-dominant}}
  \put(2.16,20.25){{\small \color[HTML]{263649}$\tilde{\alpha}^{evt}=0.12$}}
\end{picture}
\caption{MSRL objectness responses $\mathbf H_1^m$ on FRED. Dashed boxes denote ground truth. Darker shading indicates a stronger response within each map. $\tilde{\alpha}^{evt}$ is the mean Event reliability over the three scales.}
\label{fig:s-fred-responses}
\end{figure*}

\section{Additional Ablations and Hyperparameter Sensitivity}
\label{sec:s-ablations}

\subsection{Modality Degradation Protocol for RGMC}
We conduct the modality-degradation study in Table~8 of the main paper on the full five-class FRED validation set. Frames within each sequence are partitioned using a fixed random seed, yielding 75,884 image pairs per degradation condition. Before resizing to $640\times640$, we apply one degradation to each pair. RGB blur uses a $25\times25$ Gaussian kernel with $\sigma=4$, and Event dropping resets 90\% of the active pixels in the accumulated Event frame to the background value. The paired modality stays intact. Under this protocol, we evaluate AERODet with RGMC enabled and disabled.

\subsection{Parameter-Matched RCI Control}
The RGB$\times$2 control concatenates two copies of $\mathbf t^{rgb}$ and uses the same $512\to256$ projection as the joint token. Both variants have 79.99M parameters. As shown in Table~\ref{tab:s-rci-capacity}, the joint token improves AP50 by 0.8 points and mAP by 0.2 points at the same parameter count, supporting the benefit of cross-modal regional semantics. All variants calibrate objectness with the Event response.
\begin{table}[!t]
\centering\small
\setlength{\tabcolsep}{3pt}
\caption{Parameter-matched comparison of RoI tokens in RCI on five-class FRED (with Event EOC enabled).}
\label{tab:s-rci-capacity}
\begin{tabular*}{\columnwidth}{@{\extracolsep{\fill}}lcrr}
\toprule
RCI Token & Params (M) & AP50 & mAP \\
\midrule
RGB & 79.92 & 80.8 & 42.2 \\
Event & 79.92 & 80.5 & 41.9 \\
RGB$\times$2 & 79.99 & 81.3 & 42.6 \\
Joint & 79.99 & \textbf{82.1} & \textbf{42.8} \\
\bottomrule
\end{tabular*}
\end{table}

\subsection{Query Decoupling and Stop-Gradient in DSOC}
Table~\ref{tab:s-decouple-sg} ablates query decoupling and the stop-gradient operations in DSOC on the five-class FRED validation set. In the first variant, we share the final decoder query between localization and objectness while retaining the stop-gradient operations. In the second variant, we retain separate task adapters and remove the stop-gradient operations. Removing either component reduces AP50 and mAP relative to AERODet, supporting decoupled task queries and gradient isolation in DSOC.

\begin{table}[!ht]
\centering\small
\setlength{\tabcolsep}{3pt}
\caption{Ablation of query decoupling and stop-gradient in DSOC on five-class FRED.}
\label{tab:s-decouple-sg}
\begin{tabular*}{\columnwidth}{@{\extracolsep{\fill}}lrr}
\toprule
Variant & AP50 & mAP \\
\midrule
AERODet & \textbf{82.1} & \textbf{42.8} \\
w/o query decoupling & 81.1 & 42.4 \\
w/o stop-gradient & 81.3 & 42.6 \\
\bottomrule
\end{tabular*}
\end{table}

\subsection{Hyperparameter Sensitivity}
Table~\ref{tab:s-hparam} analyzes the sensitivity of AERODet to the response retention ratio $\eta$, the joint token weight $\beta$, and the objectness calibration strength $\rho$ on the five-class FRED validation set. Performance remains stable across the tested configurations, with AP50 ranging from 81.4\% to 82.2\% and mAP from 42.6\% to 43.1\%. We adopt $(\eta,\beta,\rho)=(0.01,0.1,0.5)$ as the default setting.
\begin{table}[H]
\centering\small
\setlength{\tabcolsep}{3pt}
\caption{Hyperparameter sensitivity on five-class FRED. $\dagger$ marks the default setting.}
\label{tab:s-hparam}
\begin{tabular*}{\columnwidth}{@{\extracolsep{\fill}}llrr}
\toprule
Parameter & Value & AP50 & mAP \\
\midrule
\multirow{3}{*}{$\eta$} & 0.005 & \textbf{82.2} & \textbf{43.1} \\
 & 0.01$^\dagger$ & 82.1 & 42.8 \\
 & 0.02 & 81.9 & \textbf{43.1} \\
\midrule
\multirow{3}{*}{$\beta$} & 0.1$^\dagger$ & \textbf{82.1} & \textbf{42.8} \\
 & 0.25 & 81.4 & 42.7 \\
 & 0.5 & 81.7 & 42.6 \\
\midrule
\multirow{3}{*}{$\rho$} & 0 & 81.5 & 42.7 \\
 & 0.5$^\dagger$ & \textbf{82.1} & \textbf{42.8} \\
 & 1 & \textbf{82.1} & 42.7 \\
\bottomrule
\end{tabular*}
\end{table}

\section{Additional Qualitative Visualizations}
\label{sec:s-qual}
Figure~\ref{fig:s-fred-responses} complements Figure~4 of the main paper with two more FRED cases. In (a), the Event response concentrates on the drone, while the RGB response spreads across the textured background. In (b), Event activity near the target is sparse, and the RGB response peaks on the drone. Mean Event reliability $\tilde{\alpha}^{evt}$ is 0.94 in (a) and 0.12 in (b).

\FloatBarrier

\end{document}